\documentclass[10pt,twocolumn,letterpaper]{article}

\usepackage{3dv}
\usepackage{times}
\usepackage{epsfig}
\usepackage{graphicx}
\usepackage{amsmath}
\usepackage{amssymb}

\usepackage{color}
\usepackage{xspace}
\usepackage{adjustbox}
\usepackage{tabularx,booktabs}
\usepackage{caption}

\newcommand{\norm}[1]{\left\lVert#1\right\rVert}

\usepackage{listings}
\usepackage{color}
\usepackage{capt-of}

\definecolor{dkgreen}{rgb}{0,0.6,0}
\definecolor{gray}{rgb}{0.5,0.5,0.5}
\definecolor{mauve}{rgb}{0.58,0,0.82}

\definecolor{codegreen}{rgb}{0.2,0.2,0.2}
\definecolor{codegray}{rgb}{0.5,0.5,0.5}
\definecolor{codepurple}{rgb}{0.58,0,0.82}
\definecolor{backcolour}{rgb}{0.95,0.95,0.95}

\lstdefinestyle{mystyle}{
	backgroundcolor=\color{backcolour},   
	commentstyle=\color{codegreen},
	keywordstyle=\color{magenta},
	numberstyle=\tiny\color{codegray},
	stringstyle=\color{codepurple},
	basicstyle=\footnotesize,
	breakatwhitespace=false,         
	breaklines=true,                 
	captionpos=b,                    
	keepspaces=true,                 
	numbers=left,                    
	numbersep=5pt,                  
	showspaces=false,                
	showstringspaces=false,
	showtabs=false,                  
	tabsize=2
}

\usepackage[pagebackref=true,breaklinks=true,letterpaper=true,colorlinks,bookmarks=false]{hyperref}

\threedvfinalcopy %

\def\threedvPaperID{201} %

\begin{document}

\title{\vspace{-5ex}PoseNet3D: Learning Temporally Consistent 3D Human Pose\\ via Knowledge Distillation}

\author{Shashank Tripathi$^{1^{\star}}$ \quad Siddhant Ranade$^{1,2^{\star}}$ \quad Ambrish Tyagi$^1$ \quad Amit Agrawal$^1$\\
	$^1$Amazon Lab126\quad $^2$School of Computing, University of Utah\\
\tt\small \{shatripa, ambrisht, aaagrawa\}@amazon.com, sidra@cs.utah.edu}

\maketitle

\renewcommand*{\thefootnote}{$\star$}
\setcounter{footnote}{1}
\footnotetext{Equal Contribution}

\begin{abstract}

Recovering 3D human pose from 2D joints is a highly unconstrained problem. We propose a novel neural network framework, PoseNet3D, that takes 2D joints as input and outputs 3D skeletons and SMPL body model parameters. By casting our learning approach in a student-teacher framework, we avoid using any 3D data such as paired/unpaired 3D data, motion capture sequences, depth images or multi-view images during training. We first train a teacher network that outputs 3D skeletons, using only 2D poses for training. 
The teacher network distills its knowledge to a student network that predicts 3D pose in SMPL representation. Finally, both the teacher and the student networks are jointly fine-tuned in an end-to-end manner using temporal, self-consistency and adversarial losses, improving the accuracy of each individual network. 
Results on Human3.6M dataset for 3D human pose estimation demonstrate that our approach reduces the 3D joint prediction error by $18\%$ compared to previous unsupervised methods. Qualitative results on in-the-wild datasets show that the recovered 3D poses and meshes are natural, realistic, and flow smoothly over consecutive frames.

\end{abstract}

\section{Introduction}

Accurately estimating 3D pose from 2D landmarks is a classical ill-posed problem in computer vision~\cite{forsyth2006computational,hogg1983model}. Due to projective ambiguity, there exists an infinite number of 3D poses corresponding to a given 2D skeleton~\cite{chingCVPR2019}. This makes prediction of 3D joints from 2D landmarks (lifting) a challenging task. To address these issues, previous 2D to 3D approaches have used various kinds of additional 3D supervision, including paired 2D-3D correspondences~\cite{MartinezICCV2017}, unpaired 3D data~\cite{kanazawa2018end}, multi-view images~\cite{rhodin2018learning} and synthetic data generated using motion capture (MoCap) sequences~\cite{pavlakos2018humanshape}. Acquiring MoCap data is expensive and time-consuming, and hence not scalable to new applications. Moreover, since 3D datasets do not represent all dimensions of variability in human motion, such as human shapes and sizes, appearance and clothing, environment and lighting, limb articulations etc., models trained on these datasets don't generalize to real-world scenarios~\cite{kanazawa2019learning}. In this work, we present a novel training framework, PoseNet3D, to estimate 3D human pose and shape using only 2D data as input. By starting from 2D poses, our approach allows us to train on video datasets, enabling generalization across diverse in-the-wild scenarios and emancipating us from the data-bottleneck of supervised approaches. 

Previous methods for 3D pose prediction can be classified as {model-free} and {model-based}. Typically, {model-free} approaches directly learn a mapping from 2D landmarks to 3D joints~\cite{chingCVPR2019,orinet,MartinezICCV2017}.  {Model-based} approaches fit 3D \textit{parametric models} such as SMPL~\cite{loper2015smpl} to estimate 3D shape and pose. This is typically done by minimizing the 2D error between the projection of the predicted 3D pose and the given 2D landmarks. However, as shown in~\cite{kanazawa2018end}, 2D reprojection error alone is highly under-constrained and can be minimized via non-natural joint angles. Lack of 3D supervision further aggravates this problem. 

In this paper, our goal is to train a neural network that takes 2D pose (landmarks) as input and outputs SMPL parameters and 3D skeletons, without requiring any additional 3D data or iterative fitting during training. We first train a lifting network (aka teacher) using only 2D inputs to predict \textit{model-free} 3D poses. The 3D pose output from the teacher is then used as pseudo ground truth to train a student network to predict SMPL pose parameters. Thus, our teacher-student formulation allows training the network in the absence of additional 3D data. In fact, we show that training the student network by directly minimizing the 2D reprojection error (without using knowledge from the teacher) fails due to inherent ambiguities in 2D projection, resulting in incorrect depth predictions and unnatural poses.

\begin{figure*}[tb!]
	\centering
	\includegraphics[width=0.95\textwidth,trim=0 0 0 60,clip]{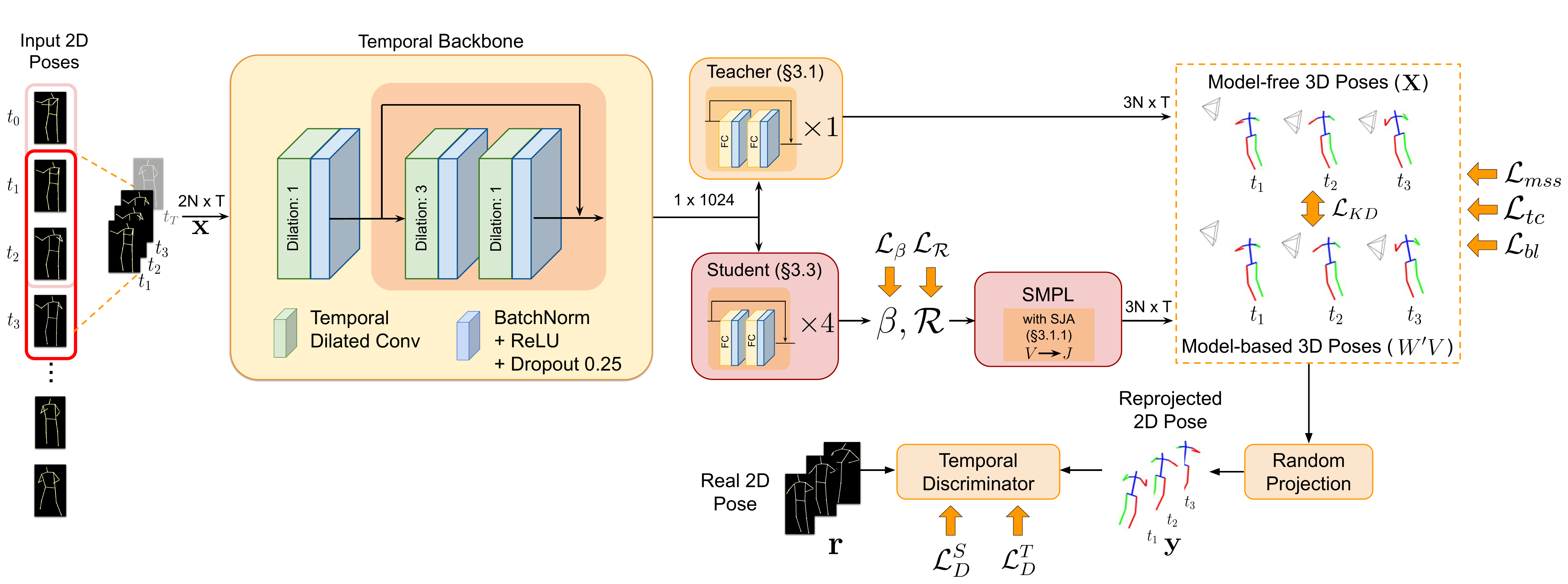}
	\vspace{-2ex}
	\caption{Overview of the proposed PoseNet3D approach. Input 2D poses are fed to a temporal backbone, followed by a teacher branch and a student branch, which output model-free 3D poses and SMPL parameters respectively.}
	\label{fig:overview}
	\vspace{-4ex}
\end{figure*}

When using a parametric model such as SMPL, there often exists a semantic gap between the SMPL 3D joints and the 2D landmarks obtained from RGB images (\eg using OpenPose~\cite{OpenPose}). For example, the 3D hip joints in SMPL are close to the center of the pelvis, while in the Human3.6M~\cite{h36m} dataset, the 2D hip joints are close to the body surface. In previous works, this semantic adaptation is learned {offline} by fitting SMPL meshes to specific 3D datasets and is used during evaluation. Thus, 3D data is also required implicitly for bridging the aforementioned semantic gap. In contrast, we demonstrate that the semantic adaptation can be automatically learned during training to bridge the gap between the SMPL 3D joints and the 2D landmarks. Bridging this semantic gap, which is often ignored in previous works is crucial, otherwise the network can minimize joint error by twisting the body, resulting in unnatural poses. 

Our approach builds upon Chen~\etal~\cite{chingCVPR2019} who train the teacher (lifter) network in an unsupervised manner. However, different from \cite{chingCVPR2019}, the primary contribution of our work demonstrates how to recover SMPL parameters from video, without requiring any 3D data or 3D pose priors for training. To the best of our knowledge, ours is the first work that shows this is feasible. It is important to note that \cite{chingCVPR2019} do not output SMPL parameters and we only use lifting as a component in our pipeline. By estimating SMPL parameters, we solve a different and arguably harder problem beyond lifting, similar to \cite{kanazawa2018end,SPIN_ICCV2019}. 

Our secondary contribution is to improve the lifting component used in our pipeline (over \cite{chingCVPR2019}), by incorporating temporal information via dilated convolutions, a temporal discriminator, and novel temporal consistency losses. 

We evaluate our approach on 3D human pose estimation tasks on Human3.6M, MPI-INF-3DHP and 3DPW datasets, reducing the mean per joint position error by $18\%$ compared to the state-of-the-art unsupervised method of~\cite{chingCVPR2019} (47mm vs 58mm) as shown in Sect.~\ref{section:evaluation}. Qualitative results confirm that our method is able to recover complex 3D pose articulations on previously unseen in-the-wild images (\eg Fig.~\ref{fig:h36vis_mesh}).

\section{Related Work}
\label{section:related_work}

Several deep learning techniques have been proposed to estimate 3D joints directly from 2D images~\cite{ChenWeaklyGeometryAwareCVPR19,ChengOcclusionAwareICCV19,HabibieInTheWildCVPR2019,IskakovLearnableTriangulationICCV2019,liang2019shapeaware,MoonICCV19,park20163d,conf/bmvc/ParkK18,Pavlakos_2017_CVPR,Rogez_2017_CVPR,SharmaMonocularICCV2019, tome2017lifting, WangNotAllPartsICCV2019, wang2019generalizing,sun2017compositional, ZhaoSemanticCVPR19,HEMletsICCV19}. 
We build upon approaches that decompose the problem into estimation of 2D joints from images followed by the estimation of 3D pose. Obtaining 2D joints from images is a mature area in itself and several approaches such as CPM~\cite{cpm}, Stacked-Hourglass (SH)~\cite{stacked-hourglass}, Mask-RCNN~\cite{mask-rcnn} or affinity models~\cite{OpenPose} can be used. 

Previous 2D to 3D approaches can be broadly classified into (a) model-free methods~\cite{CaiGCNICCV2019,ChenDeva2017,tung_2017_ICCV,IGENet,MartinezICCV2017,Moreno-Noguer_2017_CVPR,mehta2017monocular} and (b) model-based methods~\cite{kolotouros2019convolutional, kanazawa2018end, kanazawa2019learning, xu2019denserac}. Several such approaches have used 3D supervision during training. The 3D information has been used in various forms such as paired 2D-3D data~\cite{ChenDeva2017,IGENet,LiMDNCVPR19,MartinezICCV2017,mehta2017vnect,Moreno-Noguer_2017_CVPR,pavllo20193d}, 3D pose priors (\eg Gaussian Mixture Model) built using 3D MoCap sequences~\cite{bogo2016keep,SPIN_ICCV2019}, learned priors using 3D data via a discriminator~\cite{kanazawa2018end}, and synthetic 2D-3D pairings~\cite{pavlakos2018humanshape,xu2019denserac}. Our key contribution is a novel combination of model-based and model-free predictions, without requiring additional 3D data during training.

Approaches such as~\cite{chingCVPR2019,Rhodin_2018_ECCV,wang2019distill} have primarily used 2D joints from single/multi-view images without explicit 3D supervision to learn 3D pose. Chen~\etal~\cite{chingCVPR2019} proposed an unsupervised algorithm for lifting 2D poses to 3D. 
Our teacher network builds upon the work of Chen~\etal~\cite{chingCVPR2019}, but differs in the following respects. Firstly, unlike~\cite{chingCVPR2019}, our approach is able to estimate SMPL parameters. Secondly, in~\cite{chingCVPR2019}, inference uses a single frame as input and weak temporal consistency is enforced using an additional discriminator on frame differences. Their architecture only employs fully connected layers. In contrast, we use dilated convolutions (similar to~\cite{pavllo20193d}) to model temporal dynamics in the lifter as well as in the discriminator and train/test on multi-frame inputs. Video based approaches such as Li~\etal~\cite{LiBoostingICCV19} employ 3D trajectory optimization via low rank property and temporal smoothness of video sequences. Instead, we propose novel loss terms to account for the consistency of predicted skeletons on common frames across neighboring time-steps and show improvement in accuracy on the Human3.6M dataset over~\cite{chingCVPR2019}. 

\noindent\textbf{Deep Learning with SMPL:} 
Deep learning approaches such as ~\cite{arnab2019exploiting,kanazawa2018end,SPIN_ICCV2019,madadi2018smplr,omran2018neural,TexturePoseICCV19,pavlakos2018humanshape,SunHMRICCV19,varol2018bodynet} have utilized SMPL to directly regress to the underlying shape and pose parameters by training a feed-forward network. The 3D joints are computed via linear regression on the estimated mesh vertices~\cite{kanazawa2018end}. Our student network also predicts SMPL parameters but differs from these approaches in following respects. 

Firstly, approaches such as~\cite{kanazawa2018end} minimize the 2D reprojection error between the projection of the SMPL 3D joints and the predicted 2D joints from images. However, as noted in~\cite{kanazawa2018end}, 2D keypoint loss is highly unconstrained and thus~\cite{kanazawa2018end} learns the limits of joint angles using a dataset of 3D scans. Since we do not assume access to any additional 3D data at training time, we address this problem by first training a teacher network to predict 3D joint positions. We then use the output of the teacher as pseudo ground-truth to train the student network to predict SMPL parameters. By using knowledge distillation from the teacher along with simple regularizers on the SMPL parameters, we can recover realistic 3D pose without requiring additional 3D information during training. Our ablation studies show that the proposed strategy significantly outperforms the baseline strategy of directly minimizing the 2D re-projection error.

Secondly, previous works typically ignore the semantic gap between the SMPL 3D joints and the 2D landmarks while training. Instead, a regressor from vertices to joints~\cite{arnab2019exploiting,temporal3DposeURL,kanazawa2018end} is obtained offline by fitting SMPL meshes to specific 3D datasets (\eg Human3.6M). In contrast, we bridge this gap by using online \textit{semantic joint adaptation} (SJA) during training of the student network. 
We demonstrate that SJA improves the accuracy as well as naturalness of the predicted 3D pose.

\noindent\textbf{SMPL based Optimization:} Classical optimization techniques have also been used to fit the SMPL model to 2D landmarks/silhouettes~\cite{bogo2016keep,guler2019holopose,Lassner:UP:2017,mehta2017vnect}. The optimization based approaches are typically slow, prone to error and require good initialization as well as 3D pose priors built using MoCap sequences. In contrast, our approach trains a feed-forward network allowing for a faster and more robust inference. Recently, Kolotouros \etal~\cite{SPIN_ICCV2019} (\textit{SPIN}) have used iterative optimization within the training loop of a deep neural network to generate pseudo ground truth, which is used to provide direct supervision on SMPL parameters for regression. Instead, our teacher network provides supervision on 3D joints, obtained by SMPL forward kinematics. In contrast to SPIN, we do not use a 3D pose prior learned using CMU MoCap sequences (similar to~\cite{bogo2016keep}). Our approach can also be extended to use optimization in the loop to generate additional supervision on SMPL parameters, which we leave for future work.

\noindent\textbf{Knowledge Distillation:} Distilling the knowledge in neural networks has been used for applications such as network compression, combining an ensemble of models into a single model~\cite{hinton2015distilling}, enhancing privacy~\cite{papernot2017semisupervised}, large scale semi-supervised learning~\cite{fb_billionscale_2019}~\etc. Wang~\etal~\cite{wang2019distill} propose a knowledge-distillation framework to use non-rigid structure from motion (NRSfM) alogrithm as the teacher to generate pseudo ground truth for training a student network. We borrow the same terminology but our 2D-3D lifter as a teacher shows better results on Human3.6M than \cite{wang2019distill}. Unlike~\cite{wang2019distill}, our approach also outputs SMPL parameters.

\section{Proposed Approach}

Our PoseNet3D approach is a combination of model-free 3D pose estimation followed by knowledge distillation to predict SMPL pose parameters. %
As shown in Fig.~\ref{fig:overview}, the input to our network is a set of $T$ 2D skeletons from $T$ consecutive frames of a video. The architecture consists of a temporal backbone, which utilizes dilated convolutions over time to model the temporal dynamics and produces a feature vector. The feature vector is fed to two branches: (a) Teacher branch, which outputs 3D poses, and (b) Student branch, which outputs SMPL parameters. The 3D joints from the student branch are computed as described in Section~\ref{subsection:joint_estimation}. The two sets of 3D joints from the student and teacher branches are compared to ensure consistency. The predicted 3D joints from the teacher and the student branches are re-projected to 2D after random rotations and are fed to a temporal discriminator. In the following sections, we describe the teacher and student networks and associated training losses in detail.

\subsection{Teacher: Temporally Consistent Lifting}
\label{section:temporal_lifting}

Let $\mathbf{x}_i^j = (x_i^j,y_i^j), i = 1,\ldots,N$ denote the $i^{th}$ 2D pose landmark of a skeleton in frame $j$ with the root joint (mid-point between the hip joints) as origin. The 2D skeleton for frame $j$ is $\mathbf{x}^j=\left\{\mathbf{x}^j_1,\ldots,\mathbf{x}^j_N\right\}$. The input to the network at time step $t$ is a set of $T$ 2D skeleton frames of the same subject, represented as $\mathbf{x}(t)=\left\{\mathbf{x}^t,\ldots,\mathbf{x}^{t+T-1}\right\}$.
For simplicity, we drop the dependence on time-step to describe the lifter. Similar to~\cite{chingCVPR2019}, we assume a perspective camera with unit focal length centered at the origin and fix the distance of the 3D skeleton to the camera to a constant $c$ units. The 2D skeletons are normalized such that the mean distance from the head joint to the root
joint is $\frac{1}{c}$ units in 2D.

At each time-step $t$, the teacher branch predicts a depth offset $o_i^j$ for each $\mathbf{x}_i^j $. %
The 3D joints are computed as $\textbf{X}_i^j = (x_i^jz_i^j,y_i^jz_i^j,z_i^j)$, where $z_i^j=\max(1,c+o_i^j)$. The generated skeletons are projected back to 2D via random projections. Let $\mathbf{Q}$ be a random rotation matrix. The rotated 3D skeleton $\mathbf{Y}_i^j$ is obtained as
\begin{equation}
\mathbf{Y}_i^j = \mathbf{Q}(\mathbf{X}_i^j-\mathbf{X}_r^j) + \mathbf{C},
\end{equation}
where $\textbf{X}_r^j$ is the predicted root joint of $j^{th}$ skeleton and $\mathbf{C} = (0,0,c)^T$. Let $\mathbf{y}_i^j$ denote the 2D projection of $\textbf{Y}_i^j$.

\subsection{Training Losses for the Teacher Network}

\noindent\textbf{Multi-Frame Self-Supervision Loss:} Let $\mathcal{G}_T$ denote the teacher network that predicts the model-free 3D pose $\mathbf{X}_i^j = \mathcal{G}_T ( \mathbf{x}_i^j )$ as defined in Section~\ref{section:temporal_lifting}. We also lift the reprojected 2D skeletons to obtain $\widetilde{\mathbf{Y}}_i^j = \mathcal{G}_T ( \mathbf{y}_i^j )$ using the same network.
If $\mathcal{G}_T(\cdot)$ is accurate, $\widetilde{\mathbf{Y}}_i^j$ should match $\textbf{Y}_i^j$. Therefore, we define our multi-frame self-supervision loss as 
\begin{equation}
\mathcal{L}_{mss} = \sum_i^N \sum_j^T \parallel\mathbf{Y}_i^j  - \widetilde{\mathbf{Y}}_i^j \parallel^2.
\end{equation}

\noindent\textbf{Temporal Consistency Loss:} Since we predict $T$ 3D skeletons at each time step $t$, common frames exist between neighboring time-windows. Using a sliding window with temporal stride $1$, we have $T-1$ frames in common between time-step $t$ and $t+1$. We use an $\mathcal{L}_2$ loss to enforce consistency between these common frames in 3D,
\begin{equation}
\mathcal{L}_{tc} = \sum_{j=1}^{T-1}\parallel\mathbf{X}_i^{j+1}(t) - \mathbf{X}_i^j(t+1)\parallel^2.
\end{equation}

\noindent\textbf{Bone Length Loss:} At each time step $t$, we enforce that the bone lengths for the $T$ predicted 3D skeletons be consistent by minimizing the variance of bone lengths over the $T$ frames. Let $b(m,n,j) = \norm{\mathbf{X}_m^{j} - \mathbf{X}_n^{j}} $ denote the bone length between the $m^{th}$ and $n^{th}$ predicted 3D joints for frame $j$. Bone length loss is defined as
\begin{equation}
\mathcal{L}_{bl} = \sum_{m=1}^{N} \sum_{n \in \mathcal{N}(m)} \texttt{Var}_j(b(m,n,j)),
\end{equation}
where $\mathcal{N}(m)$ denotes the set of connected skeleton joints for joint $m$ and $\texttt{Var}_j$ denotes variance over $T$ frames.

\noindent\textbf{Temporal Discriminator:} The discriminator provides feedback to the lifter regarding the realism of projected 2D skeletons. In contrast to~\cite{chingCVPR2019}, which uses a single frame discriminator and a frame-difference discriminator, we use a \textit{single} temporal discriminator that takes a set of $T$ reprojected/real 2D skeletons as input. Previous approaches have used RNN and LSTM to model sequential/temporal data. A challenge in using RNN/LSTM is delayed feedback which requires the use of a policy gradient to back-propagate the feedback from the discriminator~\cite{DaiICCV2017ConditionalGAN}. In contrast, our temporal discriminator$(D)$ uses dilated convolutions and provides feedback at each time-step, simplifying the training. Formally, the discriminator is trained to distinguish between sequences of $T$ real 2D skeletons $\textbf{r}(t) = \left\{\textbf{r}^1,\ldots,\textbf{r}^T\right\}$ (target probability of $1$) and fake (projected) 2D skeletons $\textbf{y}(t) = \left\{\textbf{y}^1,\ldots,\textbf{y}^T\right\}$ (target probability of $0$). We utilize a standard adversarial loss~\cite{GAN} defined as
\begin{equation}
\small
\mathcal{L}_{D}^T = \min_{\Theta_T} \max_{\Theta_D} \mathbb{E}(\log(D(\textbf{r}(t))) +\mathbb{E}( \log(1-D(\textbf{y}(t)))),
\label{eq:adversarial_loss}
\end{equation}
where $\Theta_T$ and $\Theta_D$ denote the parameters of the teacher and the discriminator networks, respectively.

\subsection{Student: Estimating SMPL Parameters}
\label{subsection:joint_estimation}

For our model-based approach, we use the Skinned Multi-Person Linear (SMPL) representation~\cite{loper2015smpl}. SMPL is a parametric model that factors human bodies into a shape (body proportions) and pose (articulation) representation. The shape is parameterized using a PCA subspace with $300$ basis shapes and shape coefficients ($\beta$). The human pose is modeled as a set of $24$ local joint angles corresponding to $K=24$ 3D joints (including root joint) and is represented as $72$ axis-angle coefficients. We directly predict the rotation matrix corresponding to each joint from the network, on which we perform a differentiable ortho-normalization. Let $\mathcal{R}=\left\{R_1, \ldots, R_K\right\}$ denote the set of $K$ rotation matrices. Given a set of parameters $\beta$ and $\mathcal{R}$, SMPL produces a mesh $V = \mathcal{M}(\beta,\mathcal{R}), V\in \mathbb{R}^{6890\times3} $ with $6890$ vertices, where $\mathcal{M}$ is differentiable. 
Note that the 3D joints by themselves do not fully constrain the shape of the body and it is not possible to predict accurate shape using 3D joints alone. Approaches such as~\cite{pavlakos2018humanshape} have additionally used silhouettes to estimate shape and thus accurate shape prediction is not a goal of this paper. We only predict the first $10$ $\beta$ parameters (common for all $T$ frames) and set the remaining to zero. Thus, the student network has a total of $10+24\times9\times T=10+216\times T$ outputs at each time step.

\subsubsection{Semantic Joint Adaptation (SJA)}
\begin{figure}[t!]
	\centering
	\includegraphics[width=0.4\textwidth]{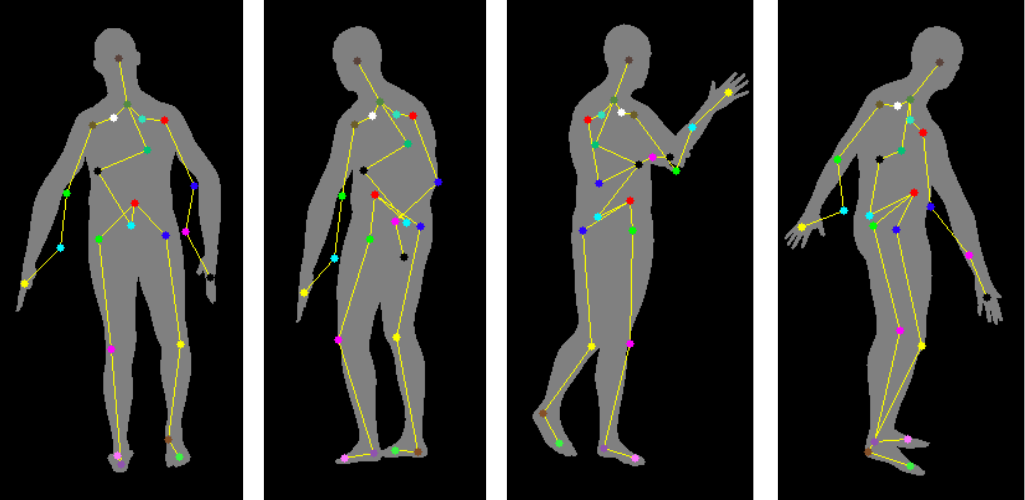}
	\vspace{-1ex}
	\caption{Linear correction on SMPL 3D joints for SJA (Equation~\ref{eq:SJA_linear}) could lead to network placing joints outside the mesh. Visualization shows rendering of predicted mesh with projection of 3D joints overlayed on top.}
	\label{fig:SJA_linear}
	\vspace{-3ex}
\end{figure}

The 3D joints $J \in \mathbb{R}^{24\times3}$ are obtained by linear regression from the final mesh vertices $V$. The linear regressor is a sparse matrix $W \in \mathbb{R}^{24\times6890}$, which represents a convex combination of vertices for each 3D joint. Hence, $J = W V$. The pre-trained linear regressor in SMPL produces 3D joints that are often semantically different from the 2D joints obtained from 2D pose detectors or annotations on datasets. For example, in SMPL the 3D hip joints are closer to the center of the body. However, in Human3.6M 2D annotations, the hip landmarks are closer to the periphery. 
Our SJA module learns the adaptation of the SMPL 3D joints to 2D joints used for training. 

\begin{figure}[tb!]
	\centering
	\includegraphics[width=0.4\textwidth]{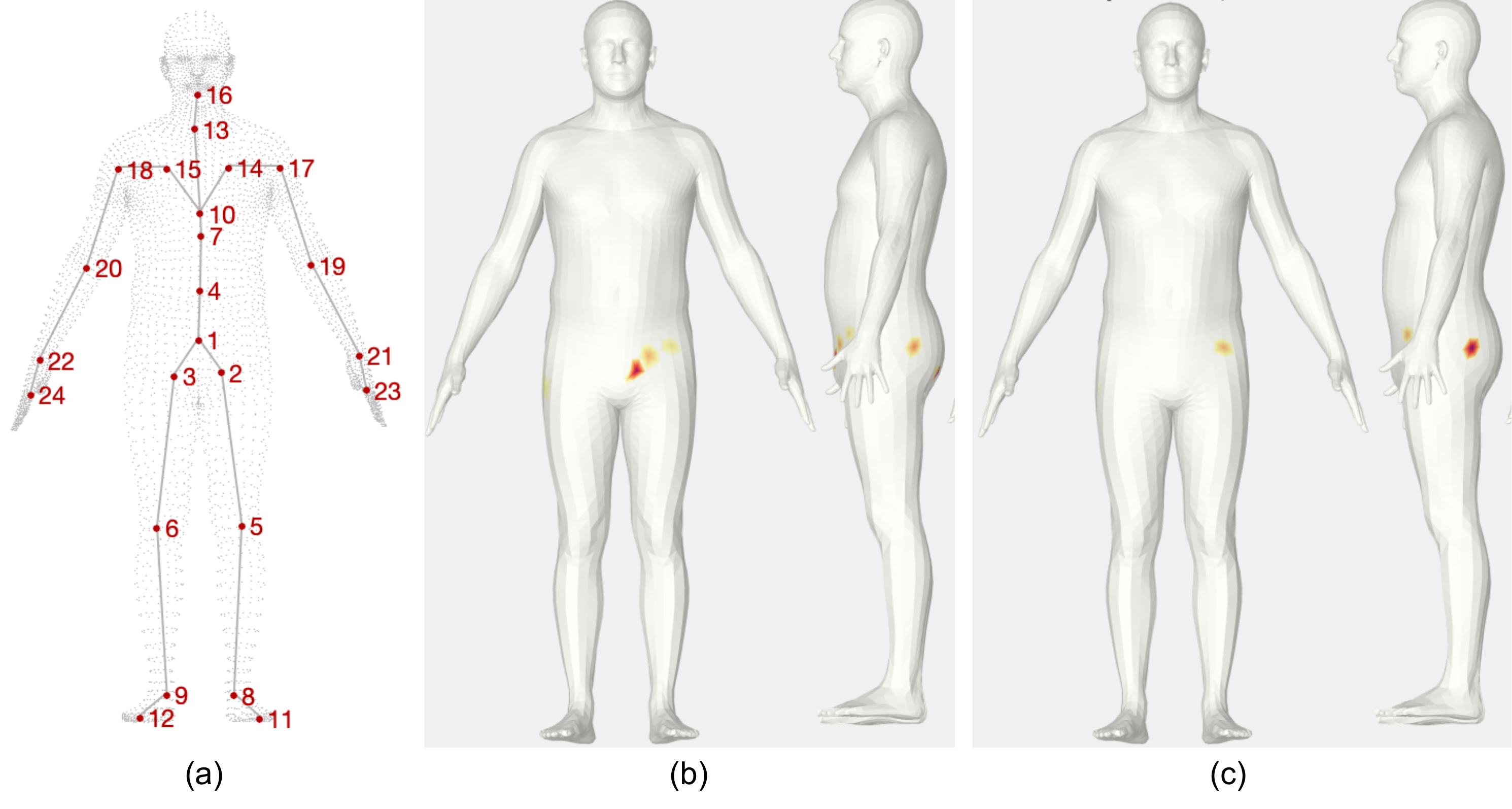}
	\vspace{-2ex}
	\caption{(a) Original SMPL 3D joints (b) SMPL regression weights for the left-hip joint \#2 are visualized by assigning a color to each vertex (dark red corresponds to higher weight) (c) Updated regressor weights for hip joint \#2 after SJA.}
	\label{fig:SJA_heatmap}
	\vspace{-3ex}
\end{figure}

We first experimented with a linear layer that learns a weight matrix $\mathbf{A} \in \mathbb{R}^{72\times72}$ and a bias vector $\mathbf{b} \in \mathbb{R}^{72\times1}$, which is applied to the $72\times 1$ vectorized representation of $J$ to adapt the SMPL joints (referred to as \textit{Linear-SJA}),
\begin{equation}
\quad \mathit{J}' = \mathbf{A} \mathit{J} + \mathbf{b}.
\label{eq:SJA_linear}
\end{equation}
However, such an approach fails in practice. Since there is no constraint on joints, the network can potentially minimize joint error by moving the SMPL joints outside the body (Fig.~\ref{fig:SJA_linear}). To avoid such pitfalls, similar to SMPL, we learn a convex combination of vertices $W'$, resulting in $24\times6890=165,360$ additional learnable parameters and obtain the new joints as $J'=W'V$. Visualization in Fig.~\ref{fig:SJA_heatmap} show that weights for the learned regressor on Human3.6M shifts from the center of body towards the surface, corresponding to a similar shift in 2D hip landmarks. For the rest of the paper, SJA refers to the learned convex combination of vertices.

\subsection{Training Losses for the  Student}

The following losses are used to train the student network via knowledge distillation.

\noindent\textbf{Knowledge Distillation Loss:} We define a loss between the model-free prediction of 3D joints $\textbf{X}_i^j$ and the 3D joints obtained via the SMPL model. To account for the mismatch between the number of joints, we choose $N$ (14 in our case) relevant joints from the 24 SMPL joints. Let $\mathcal{I}(i)$ denote the index of the SMPL joint corresponding to $i^{th}$ input 2D joint. $\mathcal{L}_{KD}$ is computed as a sum of individual losses over each joint $i$ and each frame $j$,
\begin{equation}
\mathcal{L}_{KD} = \sum_{j=1}^T \sum_{i=1}^N\parallel \textbf{X}_i^j - W'_{\mathcal{I}(i)} \mathcal{M}(\beta,\mathcal{R}^j) \parallel^2,
\end{equation}
where $W'_{\mathcal{I}(i)}$ denote the row of matrix $W'$ corresponding to the regressor weights for joint $\mathcal{I}(i)$ and $\mathcal{R}^j$ denotes the set of predicted rotation matrices for frame $j$.

\noindent\textbf{Regularization of SMPL Parameters:} In absence of any 3D data, we use a simple regularizer for pose parameters to avoid over-twisting by penalizing the deviation of the predicted rotation matrices from identity rotation.
\begin{equation}
\mathcal{L}_{R} = \sum_{j=1}^T\sum_{i=1}^K \parallel R_i^j - I_{3\times3} \parallel^2,
\end{equation}
where $I_{3\times3}$ is the $3\times3$ identity matrix. We use a similar $\mathcal{L}_2$ regularizer for $\beta$, $\mathcal{L}_{\beta} =  \parallel\beta\parallel^2$, since $\beta=0$ represents the average human shape. The $\beta$ regularizer is used with a relatively larger weight during training to keep the shape close to the average shape. %
However, we show in Section~\ref{subsec:ablation_student} (Fig.~\ref{fig:sja_no_sja_2d_repro_ablation_combined}(a)) that without SJA, these regularizers by themselves are not sufficient to avoid unnatural predictions. Our novel SJA module helps improve the realism and naturalness of predicted pose parameters.

\noindent\textbf{Discriminator:}
Similar to the teacher network, the predicted 3D joints from the student network are reprojected to random views and fed to the discriminator. The corresponding discriminator loss is $\mathcal{L}_{D}^S$, similar to $\mathcal{L}_{D}^T$ in Eqn.~\ref{eq:adversarial_loss}.

\subsection{Training PoseNet3D}
\label{section:staggered_training}

We train PoseNet3D following these steps:
\begin{enumerate}
	\itemsep0em
	\item Train Teacher: Train the shared temporal convolution backbone and the teacher branch by minimizing $\mathcal{L}_{T}=\lambda_{mss}\mathcal{L}_{mss} + \lambda_{tc}\mathcal{L}_{tc} + \lambda_{bl} \mathcal{L}_{bl} + \mathcal{L}_{D}^T $.
	\item Knowledge Distillation: Freeze the shared temporal backbone and the teacher branch. Train the student branch by minimizing $\mathcal{L}_{S}=\mathcal{L}_{KD} + \lambda_R \mathcal{L}_{R} + \lambda_{\beta} \mathcal{L}_{\beta}$.
	\item Learn SJA: Initialize $W'$ to $W$. Fine-tune $W'$ and the student branch by minimizing $\mathcal{L}_{S}$.
	\item Fine-tune the entire network by minimizing $\mathcal{L}= \mathcal{L}_{T} + \lambda_S \mathcal{L}_{S} + \mathcal{L_{D}^S}$.
\end{enumerate}
Hyper-parameters $\lambda_{mss}, \lambda_{tc}, \lambda_{bl}, \lambda_R, \lambda_{\beta}, \lambda_S$ are defined in Sect.~\ref{subsec:implementation}. Note that in step 4, we feed the re-projection of the 3D pose predicted from both the teacher and student networks to the discriminator.

\begin{figure*}[t!]
	\centering
	\includegraphics[width=0.95\textwidth]{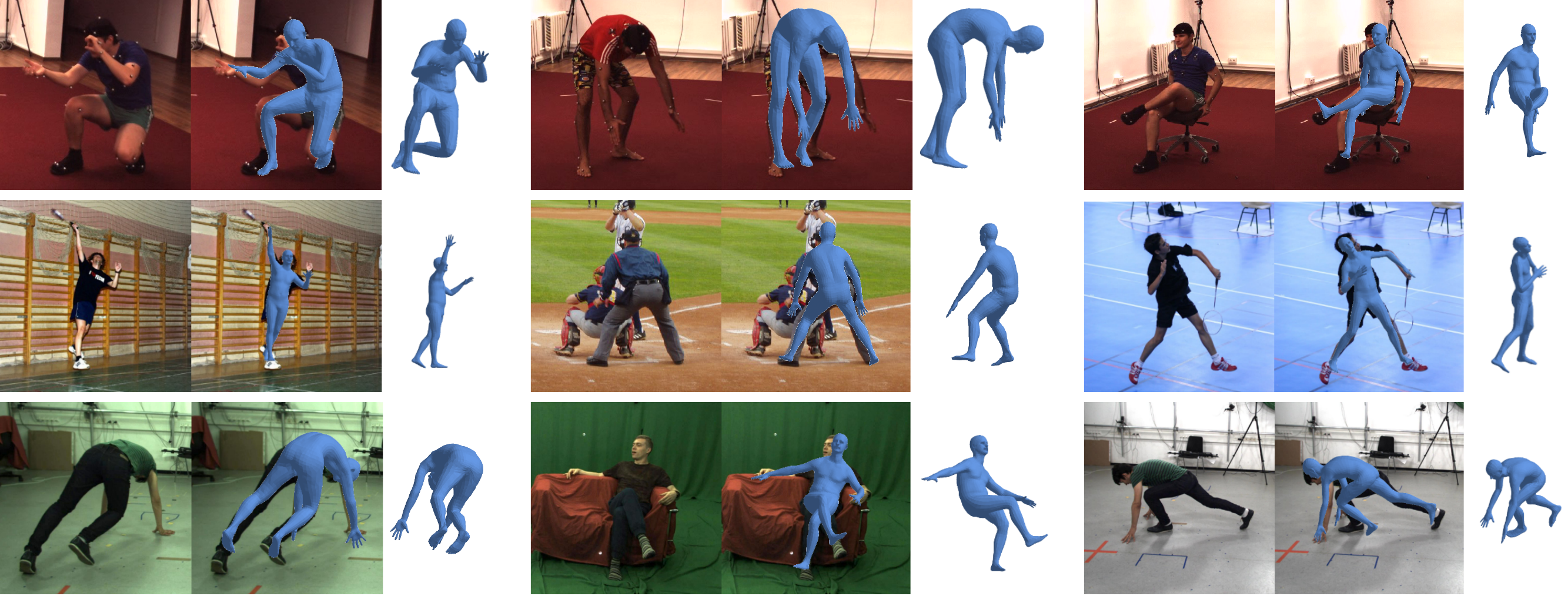} 
	\vspace{-1ex}
	\caption{Visualization of SMPL mesh obtained using predicted parameters on challenging examples. Each example shows input image, recovered SMPL mesh, and the same mesh from a different view. The student network is able to recover complicated articulations of the human body. First row: H3.6M. Second row: LSP. Third row: MPI-INF-3DHP. }
	\label{fig:h36vis_mesh}
	\vspace{-3ex}
\end{figure*}

\section{Experiments and Results}
\label{section:evaluation}
We evaluate on the widely used Human3.6M~\cite{ionescu2013human3}, MPI-INF-3DHP~\cite{mehta2017monocular} and 3DPW~\cite{vonMarcard2018}  datasets and show quantitative and qualitative results. We also show qualitative visualizations of reconstructed skeletons and meshes on in-the-wild datasets such as Leeds Sports Pose~\cite{johnson2010clustered} (LSP) where ground-truth 3D data is not available. Since our approach takes temporal sequences, for inference on single frame input (\eg LSP), we simply copy the frame $T$ times.
\subsection{Implementation Details}
\label{subsec:implementation}
We use $N=14$ joints and randomly sample $T=9$ frame sub-sequences of 2D poses from videos for training. The input poses are normalized such that the mean distance from the head joint to the root joint is $0.1$ units, corresponding to placing the 3D skeletons at $c=10$ units from the camera.
As shown in Fig.~\ref{fig:overview}, our temporal backbone takes a $2N \times T$ input followed by a \textit{conv-block}, comprising convolution filter, batchnorm, ReLU and dropout. Each convolution filter has $ 1024 $ channels with a kernel size of $ 3 \times 1 $ and temporal dilation factor of $ d=1 $.  The output of the \textit{conv-block} is fed to a residual block with two \textit{conv-blocks}, with a dilation ratio of $d=3$ and $d=1$, respectively. The teacher branch consists of an additional residual block with $2$ fully-connected (FC) layers of size $1024$ each. Similarly, the student branch consists of $4$ FC residual blocks. The temporal discriminator architecture is identical to the temporal backbone architecture but does not use BatchNorm. We train on TitanX GPUs using the Adam optimizer~\cite{kingma2014adam} with batchsize of $6000$ and learning rate of $0.0001$ for 150 epochs. The loss weights are empirically set as $\lambda_{mss}$= $2$, $\lambda_{tc}$= $1$, $\lambda_{bl}$= $2$, $\lambda_R$= $30$, $\lambda_{\beta}$= $10$ and $\lambda_S$= $2$.

\subsection{Datasets and Metrics}

\noindent\textbf{Human3.6M (H3.6M):} This is one of the largest 3D human pose datasets, consisting of 3.6 million 3D human poses. The dataset contains video and MoCap data from 11 subjects performing typical activities such as walking, sitting,~\etc. Similar to previous works~\cite{chingCVPR2019,ZedNet_2018_ECCVW,kanazawa2018end}, we report the mean per joint position error in mm after scaling and rigid alignment to the ground truth skeleton (P-MPJPE) on subjects S9 and S11 (all cameras).
We only use 2D data from subjects S1, S5, S6, S7 and S8 for training a single activity-agnostic model. To evaluate the smoothness of predicted 3D pose, we %
report mean per joint velocity error (MPJVE), which is calculated as the mean per joint error of the first derivative of the aligned 3D pose sequences (in mm/frame at 50 Hz). We also propose the mean bone-length standard deviation (MBLSTD) metric as the average standard deviation (in mm) of 8 bone segments (corresponding to upper/lower/left/right hand/leg) over all sequences. Lower values of MPJPE, MPJVE, and MBLSTD signify better performance.

\noindent\textbf{MPI-INF-3DHP:}
The MPI-INF-3DHP dataset consists of 3D data captured using a markerless MoCap system. We evaluate on valid images from test-set containing 2929 frames following~\cite{kanazawa2018end} and report P-MPJPE, Percentage of Correct Keypoints (PCK) $@$150mm, and Area Under the Curve (AUC) computed for a range of PCK thesholds. 

\noindent\textbf{3DPW:}
3DPW~\cite{vonMarcard2018} is a recent outdoor 3D dataset containing 60 videos wherein 3D ground-truth annotations are extracted using IMUs attached to body limbs. Similar to \cite{SPIN_ICCV2019}, we only use this dataset for evaluation. To handle missing joints, we follow \cite{chingCVPR2019} and train a supervised joint filler.

\subsection{Quantitative Results}

\begin{table}[tb!]
	\centering
	\small
	\begin{adjustbox}{max width=0.99\textwidth}
	\begin{tabular}{llcc}
		\toprule
		Supervision & Method & \multicolumn{2}{c}{P-MPJPE} \\
		&   & GT & IMG\\
		\midrule 
		Full & Chen and Ramanan~\cite{ChenDeva2017}  & 57.5 & 82.7 \\
    	& Martinez~\etal~\cite{MartinezICCV2017}  & 37.1 & 52.1\\		
		& IGE-Net~\cite{IGENet} (17j)  & 35.8 & 47.9\\
		& Li and Lee~\cite{LiMDNCVPR19} &  - & 42.6\\
		& Ci~\etal~\cite{CiICCV2019GCN}    & 27.9 & 42.2\\
		& Hossain \& Little~\cite{rayat2018exploiting} (17j)  $\dagger$& & 42.0\\
		& Pavllo \etal~\cite{pavllo20193d} $\dagger$   & 22.7 & 40.1\\
		& Cai~\etal~\cite{CaiGCNICCV2019} $\dagger$   & - & 39.0\\
		& Yang~\etal~\cite{Yang_2018_CVPR}(+) & - & 37.7\\

		\midrule
		Weak/Self & 3DInterpreter~\cite{InterpreterNetwork2016}   & 88.6 & 98.4\\
		& Tung~\etal~\cite{tung_NIPS_2017}    & - & 98.4\\
		& AIGN~\cite{tung_2017_ICCV}    & 79.0 & 97.2\\
		& RepNet~\cite{RepNet} & 38.2 & 65.1\\
		& Drover~\etal~\cite{ZedNet_2018_ECCVW} & 38.2 & 64.6 \\
		&Wang~\etal~\cite{wang2019distill}   & - & 62.8\\
		& Kocabas~\etal~\cite{Kocabas_CVPR_2019}($\ddag$) & - & 60.2 \\
		
		\midrule
		Unsupervised & Rhodin~\etal~\cite{Rhodin_2018_ECCV}($\ddag$)  & - & 98.2 \\
		&Chen~\etal~\cite{chingCVPR2019} & 58.0 & -\\
		&Chen~\etal~\cite{chingCVPR2019}$(\dagger)$(+) & 51.0 & 68.0\\
		&PoseNet3D-Teacher$(\dagger)$   & 50.6 & 66.6 \\		%
		&PoseNet3D-Teacher-FT$(\dagger)$  &\textbf{46.7} &  \underline{62.1}\\ 		%
		&PoseNet3D$(\dagger)$  &\underline{47.0} &  \textbf{59.4}\\ 		%
		
			\midrule
		Semi-supervised  & Chen~\etal~\cite{chingCVPR2019}  & 37 & - \\
		(5\% 3D data) & PoseNet3D-Teacher$(\dagger)$ & 35 & 57.1 \\
		&PoseNet3D$(\dagger)$  & 33.7 &  53.4\\ 		%

		\bottomrule
	\end{tabular}
	\end{adjustbox}
	\caption{\textbf{Human3.6M}. Comparison of P-MPJPE (in mm) for model-free 3D pose estimation. GT and IMG denote results obtained using ground truth 2D annotations and estimated 2D pose by SH/CPM~\cite{stacked-hourglass,cpm} respectively. Best and second best results are bolded and underlined, respectively. $(\dagger)$ using temporal information, ($\ddag$) using multi-view data. (17j) using 17 joints. (+) using additional data for training.}
	\vspace{-3ex}
	\label{table:h36_teacher}
\end{table}

\begin{table}[tb!]
	\centering
	\small
	\begin{adjustbox}{max width=0.95\linewidth}
		\begin{tabular}{lll}
			\toprule
			Method & 3D Data (Training) &  P-MPJPE \\
			
			\midrule
			NBF~\cite{omran2018neural}  & Paired &   59.9  \\
			HMR~\cite{kanazawa2018end} (All/Frontal Cam)  & Paired &   58.1/56.8  \\
			HMR-Video~\cite{kanazawa2019learning}$\dagger$  & Paired &   57.8  \\		 
			DenseRaC~\cite{xu2019denserac}  & Paired &   51.4  \\	
			Kolotouros~\etal~\cite{kolotouros2019convolutional}  & Paired &   50.1 \\		 	
			DenseRaC~\cite{xu2019denserac}  & Synthetic  &   48.0  \\ 
			HoloPose~\cite{guler2019holopose}  & Paired &  46.5  \\
			Sun~\etal~\cite{SunHMRICCV19}  & Paired &  42.4  \\
			
			SPIN~\cite{SPIN_ICCV2019} & Paired  &   41.1\\
			\midrule
			HMR~\cite{kanazawa2018end}(All/Frontal Cam) & Unpaired &   67.5/66.5  \\
			SPIN~\cite{SPIN_ICCV2019} & CMU Pose Prior  &   {62.0} \\

			\midrule
			PoseNet3D-Student ($\dagger$)  & None &  63.7 \\
			PoseNet3D-Student-FT ($\dagger$) & None &  \underline{60.5}  \\
			PoseNet3D ($\dagger$) & None &  \textbf{59.4} \\		 %
			\bottomrule
		\end{tabular}
	\end{adjustbox}
	\vspace{-1ex}
	\caption{\textbf{Human3.6M}. Comparison of our student network with previous approaches that output SMPL parameters (in mm). Best and second best results are bolded and underlined, respectively. Our results use SH~\cite{stacked-hourglass} for 2D pose inputs. \textit{3D data} refers to the use of additional 3D data during training. $(\dagger)$: using temporal information.}
	\label{table:h36_student}
	\vspace{-2ex}
\end{table}

\begin{table}[tb!]
	\centering
	\small
	\begin{adjustbox}{max width=0.55\linewidth}
			\begin{tabular}{lcc}%
			\toprule
			Method & P-MPJPE \\
			\midrule 
			HMR~\cite{kanazawa2018end}  & 81.3   \\
			Doersch~\etal~\cite{doersch2019sim2real}   &  74.7   \\
			HMR-Video~\cite{kanazawa2019learning}   & 72.6    \\
			Arnab~\etal~\cite{arnab2019exploiting} &  72.2   \\
			Kolotouros~\etal~\cite{kolotouros2019convolutional}   &  70.2   \\
			Sun~\etal~\cite{sun2019human}   &  69.5   \\
			SPIN~\cite{SPIN_ICCV2019} - static fits   &  66.3   \\
			SPIN~\cite{SPIN_ICCV2019} - in the loop  & 59.2 \\
			\midrule
			PoseNet3D (IMG) & 73.6 \\
			PoseNet3D (GT) & 63.2 \\
			\bottomrule
		\end{tabular}%
	\end{adjustbox}
	\vspace{-1ex}
	\caption{\textbf{3DPW}.  Comparison with previous apporaches that output SMPL parameters (in mm). Unlike other approaches, our approach does not use any 3D data}
	\label{table:3dpw_results}
	\vspace{-2ex}
\end{table}

We denote our results obtained by taking the average of the predicted 3D poses from the teacher and the student networks as \textit{PoseNet3D}. Averaging dampens output noise and improves upon both branches (Table~\ref{table:h36_teacher},~\ref{table:h36_student}), thereby showing an implicit ensemble effect of the two branches. For a fair comparison with model-free methods in Table~\ref{table:h36_teacher}, we also report the results from the teacher branch (\textit{PoseNet3D-Teacher}) after Step $1$ of training (Sect.~\ref{section:staggered_training}). Similarly, results from the student branch after Step $3$ of training are denoted as \textit{PoseNet3D-Student} in Table~\ref{table:h36_student}. The corresponding results from the two branches after fine tuning (Step $4$, Sect.~\ref{section:staggered_training}) are denoted by \textit{PoseNet3D-Teacher-FT} and \textit{PoseNet3D-Student-FT}, respectively. 

As evident from Table~\ref{table:h36_teacher}, all variations of our approach trained using H3.6M data outperform the state-of-the-art unsupervised algorithm of Chen~\etal~\cite{chingCVPR2019} (trained using H3.6M only). Our best model reduces the P-MPJPE error from $58$mm to $47$mm ($18\%$ improvement). We also outperform several previous weakly-supervised approaches that use 3D information in training and with just 5\% 3D data, our approach performs competitively with fully supervised approaches.  Similarly, on H3.6M, our results are better than previous model-based approaches such as HMR and SPIN that use unpaired 3D data and produce SMPL meshes as output (Table~\ref{table:h36_student}). On in-the-wild 3DPW dataset, we show comparable performance to many recent approaches that use 3D data or 3D pose priors (Table~\ref{table:3dpw_results}). Finally, Table~\ref{table:MPI} summarizes results on MPI-INF-3DHP. \textit{PoseNet3D} model trained on H3.6M outperforms HMR~\cite{kanazawa2018end} and comes close to the results from SPIN~\cite{SPIN_ICCV2019}, both of which were trained on MPI-INF-3DHP and used unpaired 3D data for training. This offers a strong evidence that \textit{PoseNet3D} generalizes well to out-of-domain datasets (\eg, in this case trained on H3.6M and tested on MPI-INF-3DHP).

\begin{table}[tb!]
	\centering
	\begin{adjustbox}{max width=0.99\linewidth}
		\begin{tabular}{lllccc}%
			\toprule
			Method & 3D Data & Training Datasets & \multicolumn{3}{c}{Rigid Alignment}  \\
						& (for Training) &  & & &  \\
			& & & PCK & AUC & P-MPJPE \\
			\midrule 
     		Vnect~\cite{mehta2017vnect} & Paired & H3.6M+MPI-INF-3DHP & 83.9 & 47.3 & 98.0 \\
			HMR~\cite{kanazawa2018end} & Paired & H3.6M+MPI-INF-3DHP & 86.3 & 47.8 & 89.8\\
			DenseRaC~\cite{xu2019denserac} &  Paired+Unpaired & Synthetic+Various & 89.0 & 49.1  & 83.5\\
			SPIN~\cite{SPIN_ICCV2019} &  Paired & Various & 92.5 & 55.6 & 67.5\\

			\midrule
			 HMR~\cite{kanazawa2018end}& Unpaired & H3.6M+MPI-INF-3DHP & 77.1 & 40.7 & 113.2\\ 
			SPIN~\cite{SPIN_ICCV2019} & Unpaired & Various & {87.0} & {48.5} & {80.4 } \\ 
			\midrule
			PoseNet3D & None & H3.6M & {81.9} & {43.2} & {102.4}\\			
			\bottomrule
		\end{tabular}%
	\end{adjustbox}
	\vspace{-1ex}
	\caption{\textbf{MPI-INF-3DHP}. Comparison with previous approaches that output SMPL parameters. Metrics for~\cite{kanazawa2018end,SPIN_ICCV2019,mehta2017vnect} are taken from~\cite{SPIN_ICCV2019}. \textit{Various} refers to combination of datasets such as H3.6M, MPI-INF-3DHP and LSP. PCK and AUC: higher is better. P-MPJPE (mm): lower  is better. }
	\vspace{-3ex}
	\label{table:MPI}
\end{table}

\subsection{Qualitative Results}

\begin{figure}[htb]
	\centering
	\includegraphics[width=0.455\textwidth]{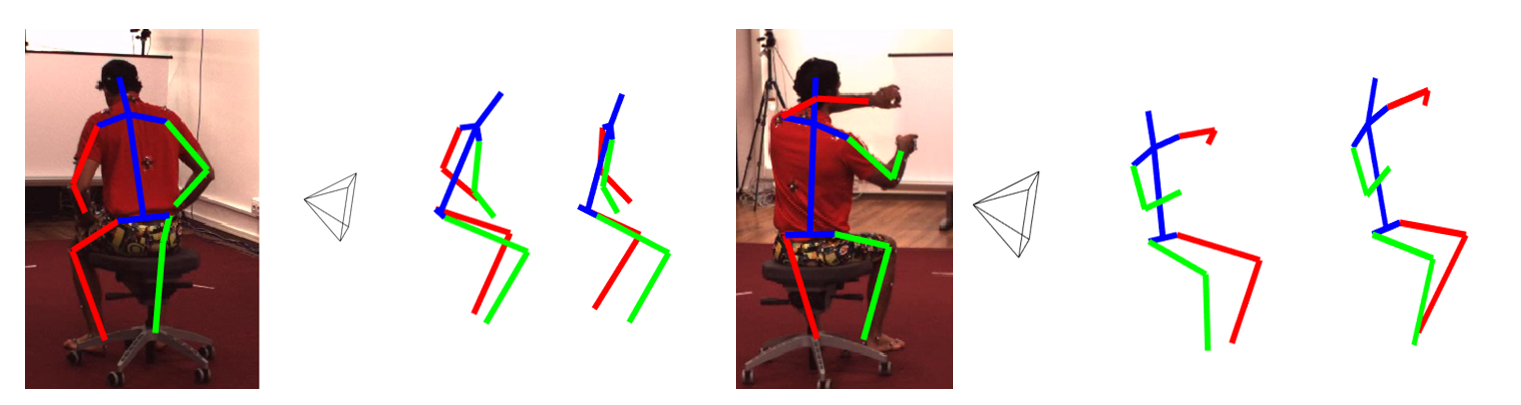} \\
	\includegraphics[width=0.385\textwidth]{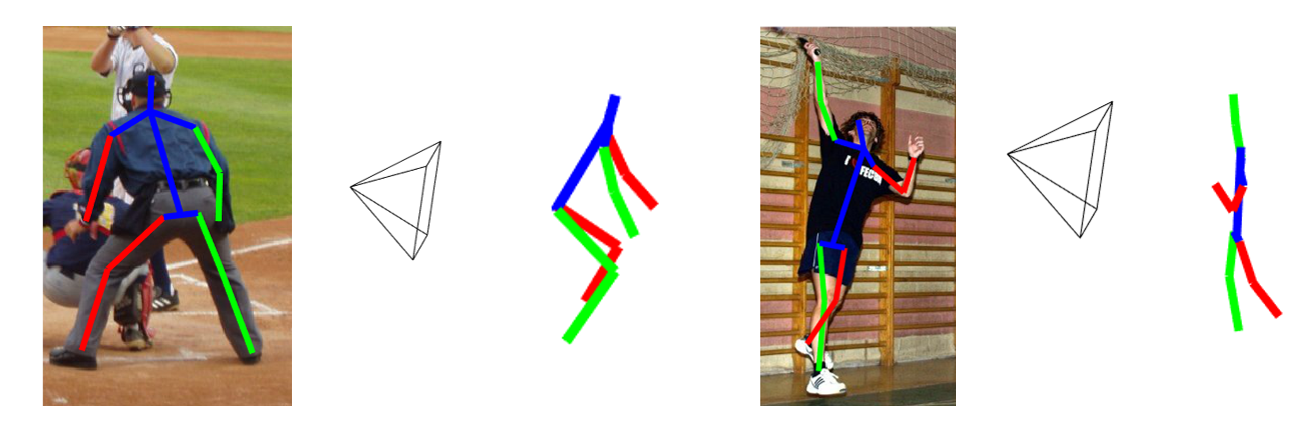}
	\vspace{-2ex}
	\caption{Visualization of predicted 3D pose on H3.6M (top) and LSP (bottom). For H3.6M, the first skeleton in each example shows ground-truth 3D skeleton.}
	\label{fig:h36vis_skeleton}
	\vspace{-3ex}
\end{figure}

Figure~\ref{fig:h36vis_mesh} shows overlay of generated mesh using predicted SMPL parameters on the corresponding image, for a few examples from H3.6M, LSP and 3DHP datasets. 
As discussed earlier, since our approach uses only 2D landmarks and cannot estimate accurate shape, the projected mesh may not align well with the human silhouette in the image. However, note that our approach is able to recover complicated articulations of the human body. Fig.~\ref{fig:h36vis_skeleton} shows predicted 3D skeletons from the teacher network on examples from H3.6M and LSP datasets. Finally, Fig.~\ref{fig:h36vis_mesh_failure} presents a few failure cases of our approach from H3.6M dataset. Please see supplementary material for additional examples.

\begin{figure}
	\centering
	\includegraphics[width=0.475\textwidth]{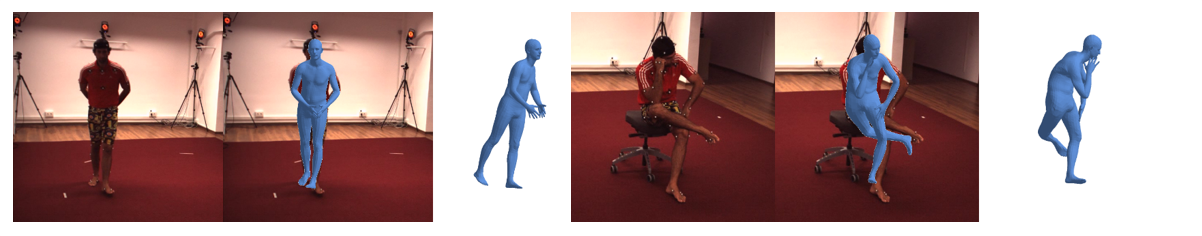} \\
	\vspace{-1ex}
	\caption{Some failure examples from Human3.6M, depicting front/back depth ambiguity in predicting joints.}
	\label{fig:h36vis_mesh_failure}
	\vspace{-3ex}
\end{figure}

\subsection{Ablation Studies}
\label{subsec:ablation_student}

\begin{table}[tb!]
	\centering
	\setlength{\tabcolsep}{2pt}
	\begin{adjustbox}{max width=0.99\linewidth}
		\begin{tabular}{lcc||lc}%
			\toprule
			Ablation & MPJVE & MBLSTD & Student Ablation & P-MPJPE\\
			\midrule 
			\cite{chingCVPR2019}$\ddag$ (1 frame)  & 22.2 & 43.7 & 	Baseline 1 &  124.8  \\
			\cite{chingCVPR2019}$\ddag$ (3 frames)  & 21.0 & 38.7 & 	Baseline 2 &  92.5   \\
			
			PoseNet3D-Teacher (1 frame) &  23.4 & 41.6 & 	PoseNet3D-S-LinearSJA  & 108.4    \\
			PoseNet3D-Teacher (9 frames) &  {13.9} &{40.1} & 	PoseNet3D-S-NoSJA  & 70.2  \\
			PoseNet3D-Teacher-FT (9 frames) &  {13.6} & {38.7} & 	PoseNet3D-S-SJA    & \textbf{63.7}\\
			PoseNet3D (9 frames) &  \textbf{7.8} & \textbf{27.0} \\
			
			\bottomrule
		\end{tabular}%
	\end{adjustbox}
	\vspace{-1ex}
	\caption{Ablation studies. (Left) Temporal consistency. (Right) Effect of SJA on student network. Reported metrics use 2D joints obtained from SH~\cite{stacked-hourglass}. $\ddag$ denotes our implementation of~\cite{chingCVPR2019}.
	}
	\label{table:combined-ablation}
	\vspace{-3ex}
\end{table}

Table~\ref{table:combined-ablation} analyzes the impact of number of frames for the teacher network with and without fine-tuning in terms of MPJVE and MBLSTD. We implemented the approach of \cite{chingCVPR2019} to compute similar metrics and our $9$-frame teacher network outperforms their approach, reducing MPJVE and MBLSTD by more than $60\%$ and $30\%$, respectively.

\begin{figure}[t!]
		\centering
		\includegraphics[width=0.99\linewidth]{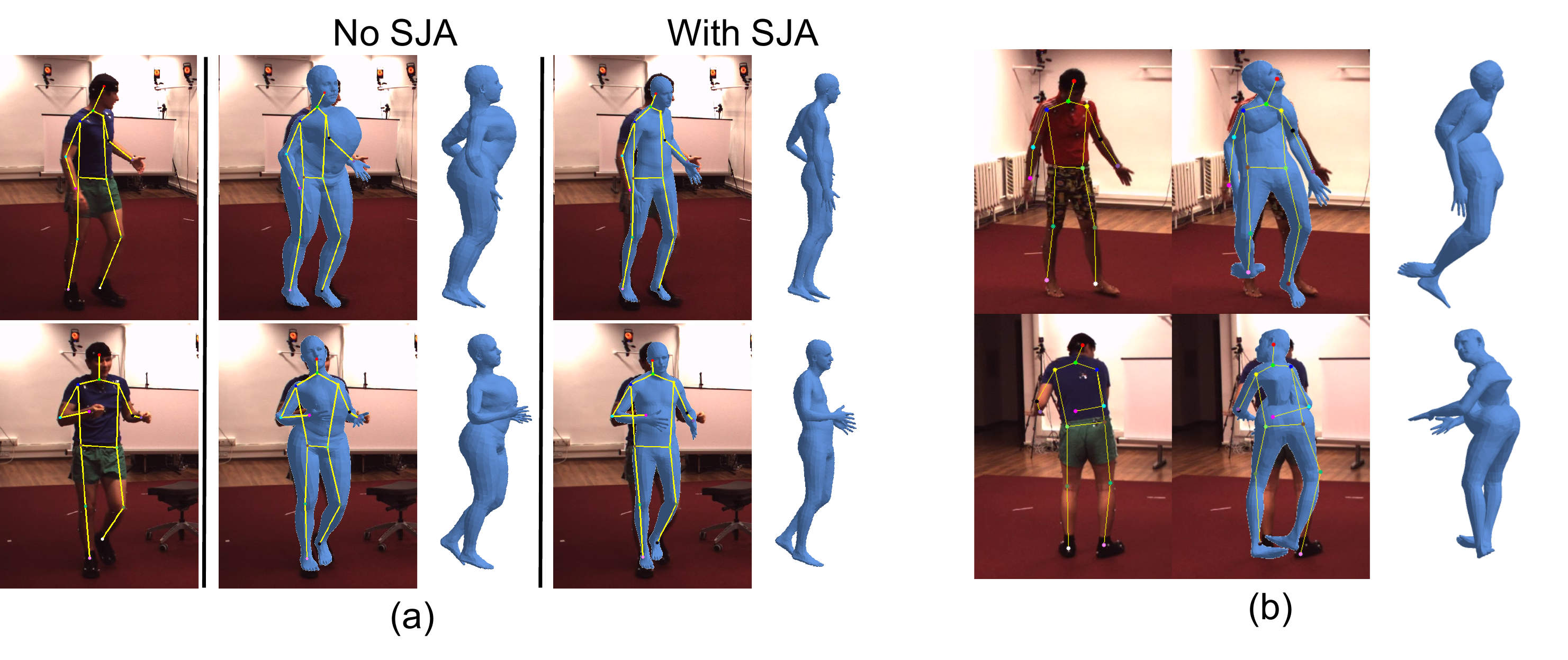}
		\vspace{-2ex}
		\caption{(a) Effect of SJA on the student network. (b) Without knowledge distillation, directly training the student network with 2D re-projection loss results in \textit{monster} meshes, even when the 2D loss is small. 
		}
		\label{fig:sja_no_sja_2d_repro_ablation_combined}
	\vspace{-3ex}
\end{figure}

For the student branch, we first define two baselines. Since SMPL is a parametric model, a trivial baseline is to train the student network directly by minimizing the 2D re-projection error (\textit{Baseline 1}). We also propose an additional baseline by employing SJA on top (\textit{Baseline 2}). As noted in~\cite{arnab2019exploiting} and specifically in~\cite{kanazawa2018end}, minimizing the 2D re-projection error without any 3D supervision can result in \textit{monster} meshes with high P-MPJPE. We observe a similar phenomenon. As shown in Table~\ref{table:combined-ablation} and Fig.~\ref{fig:sja_no_sja_2d_repro_ablation_combined}(b), these baselines result in a high P-MPJPE and do not predict high quality poses.

Finally, we analyze the effect of SJA on the student network using KD. Linear-SJA (Eqn.~\ref{eq:SJA_linear}) results in low 3D error on training data but high 3D error on test data (\textit{PoseNet3D-S-LinearSJA}). The network severely overfits on the training data by moving joints outside the body (see Fig.~\ref{fig:SJA_linear} for examples). Without using SJA (\textit{PoseNet3D-S-noSJA}), the error is higher than using SJA (\textit{PoseNet3D-S-SJA}). Visualization in Fig.~\ref{fig:sja_no_sja_2d_repro_ablation_combined}(a) compares the output of student network with and without SJA. Notice how the re-projected 3D skeleton is semantically closer to the input 2D skeleton with SJA (especially hip and head joints). SJA results in better 3D pose predictions confirming that in absence of any paired/unpaired 3D supervision, our semantic joint adaptation module is essential for training the student network.

\section{Conclusions}
\label{section:conclusion}
We present a knowledge distillation algorithm to learn SMPL pose parameters from 2D joints, without requiring additional 3D data for training. Our approach trains a feed-forward network to predict SMPL parameters and does not require any iterative fitting. We first learn a teacher network to lift 2D joints to model-free 3D pose in a temporally consistent manner. The temporal dynamics are modeled using dilated convolutions in both lifter and discriminator, allowing feedback at every time-step and avoids common pitfalls in using LSTM/RNN in such settings. The teacher network provides pseudo ground truth to the student network which learns to predict SMPL pose parameters. We demonstrate how to bridge the semantic gap between the SMPL 3D joints and 2D pose landmarks during training, which has been largely ignored in previous literature. We believe that our paper has significantly improved the state-of-art in learning of 3D pose from 2D skeletons in absence of explicit 3D supervision.

\clearpage

{\small
\bibliographystyle{ieee}
\bibliography{egbib,poseRef,ching,references}
}
\vspace{115ex}
\begin{figure*}[!htb]
	\centering
	\textbf{\large Supplementary Material\\PoseNet3D: Learning Temporally Consistent 3D Human Pose\\ via Knowledge Distillation}
\end{figure*}
\pagebreak

\appendix

\section{Implementation Details}

Our pipeline is implemented in python using the PyTorch deep-learning library. Training takes 30 hrs on 4 Nvidia TitanX GPUs. Inference time, given 2D pose input, is 20ms on the same GPU. In the following subsections, we furnish relevant empirical details for our experiments in the manuscript. \\

\noindent \textbf{Architecture of the Temporal Generator.}
The generator network consists of a shared temporal backbone, followed by a model-free teacher branch and a model-based student branch. We attach the pytorch model dump of the generator below.

\begin{lstlisting}
(TemporalBackBone): 
(0):  Conv1d(28, 1024, ksz=3, st=1, dil=1)
(1):  BatchNorm1d(1024, eps=1e-05, mmntm=0.1)
(2):  ReLU(inplace)
(3):  Dropout(p=0.25)
ModuleList(
(4):  Conv1d(1024, 1024, ksz=3, st=1, dil=3)
(5):  BatchNorm1d(1024, eps=1e-05, mmntm=0.1)
(6):  ReLU(inplace)
(7):  Dropout(p=0.25)
(8):  Conv1d(1024, 1024, ksz=3, st=1, dil=1)
(9):  BatchNorm1d(1024, eps=1e-05, mmntm=0.1)
(10): ReLU(inplace)
(11): Dropout(p=0.25))
		  Shortcut(res+in(4))
\end{lstlisting}

\begin{lstlisting}
(TeacherBranch): 
ModuleList(
(0): Linear(in_f=1024, out_f=1024, bias=True)
(1): BatchNorm1d(1024, eps=1e-05, mmntm=0.1)
(2): ReLU(inplace)
(3): Linear(in_f=1024, out_f=1024, bias=True)
(4): BatchNorm1d(1024, eps=1e-05, mmntm=0.1)
(5): ReLU(inplace))
		 Shortcut(res+in(0))
(6): Linear(in_f=1024, out_f=14, bias=True)
\end{lstlisting}

\begin{lstlisting}
(StudentBranch): 
ModuleList(
(0): Linear(in_f=1024, out_f=1024, bias=True)
(1): BatchNorm1d(1024, eps=1e-05, mmntm=0.1)
(2): ReLU(inplace)
(3): Linear(in_f=1024, out_f=1024, bias=True)
(4): BatchNorm1d(1024, eps=1e-05, mmntm=0.1)
(5): ReLU(inplace))
		 Shortcut(res+in(0))
ModuleList(
(6): Linear(in_f=1024, out_f=1024, bias=True)
(7): BatchNorm1d(1024, eps=1e-05, mmntm=0.1)
(8): ReLU(inplace)
(9): Linear(in_f=1024, out_f=1024, bias=True)
(10): BatchNorm1d(1024, eps=1e-05, mmntm=0.1)
(11): ReLU(inplace))
		  Shortcut(res+in(6))
ModuleList(
(12): Linear(in_f=1024, out_f=1024, bias=True)
(13): BatchNorm1d(1024, eps=1e-05, mmntm=0.1)
(14): ReLU(inplace)
(15): Linear(in_f=1024, out_f=1024, bias=True)
(16): BatchNorm1d(1024, eps=1e-05, mmntm=0.1)
(17): ReLU(inplace))
		  Shortcut(res+in(12))
ModuleList(
(18): Linear(in_f=1024, out_f=1024, bias=True)
(19): BatchNorm1d(1024, eps=1e-05, mmntm=0.1)
(20): ReLU(inplace)
(21): Linear(in_f=1024, out_f=1024, bias=True)
(22): BatchNorm1d(1024, eps=1e-05, mmntm=0.1)
(23): ReLU(inplace))
		  Shortcut(res+in(18))
(24): Linear(in_f=1024, out_f=14, bias=True) 
(25): Linear(in_f=1024, out_f=1944, bias=True)
(26): Linear(in_f=1024, out_f=10, bias=True)
\end{lstlisting}

The acronyms used in the pytorch model dump are described in Table.~\ref{tab:acronym}.
\begin{table}[h]
	\centering
	\begin{tabular}{r|l}
		\hline
		\textbf{Acronym} & \textbf{Meaning} \\
		\hline
		ksz & kernel\_size \\ \hline
		st & stride \\ \hline
		pdng & padding \\ \hline
		mmntm & momentum\\ \hline
		in\_f & in\_features \\ \hline
		out\_f & out\_features \\ \hline
		dil & dilation \\ \hline
		in(n) & input to layer n \\ \hline
	\end{tabular}
	\caption{Acronyms used within PyTorch model dumps.}
	\label{tab:acronym}
\end{table}

\noindent \textbf{Architecture of the Temporal Discriminator.}
The temporal discriminator architecture is identical to the temporal backbone with the exception of not having BatchNorm. The pytorch model dump is attached below: 
\begin{lstlisting}
(TemporalDiscriminator): 
(0): Conv1d(28, 1024, ksz=3, st=1, dil=1)
(1): ReLU(inplace)
(2): Dropout(p=0.25)
ModuleList(
(3): Conv1d(1024, 1024, ksz=3, st=1, dil=3)
(4): ReLU(inplace)
(5): Dropout(p=0.25)
(6): Conv1d(1024, 1024, ksz=3, st=1, dil=1)
(7): ReLU(inplace)
(8): Dropout(p=0.25))
		 Shortcut(res+in(3))
(9): Linear((in_f=1024, out_f=1, bias=True) 
\end{lstlisting}

\noindent \textbf{Camera Assumptions.}
Due to the fundamental perspective ambiguity, absolute metric depths cannot be obtained from a single view. To resolve this, we assume a camera with unit focal length centered at origin (0,0,0) and normalize the distance of the ground-truth 3D skeletons from the camera to a constant $c=10$m and a constant scale (head to root joint distance) of $1$m. We also normalize the input 2D skeletons such that the mean distance from the head joint to the root joint is $\frac{1}{c}=0.1$ units in 2D. This ensures that 3D skeletons will be generated with a constant scale of $\approx 1$ m (head to root joint distance). $z_i^j=\max(1,c+o_i^j)$ further constrains the predicted 3D skeleton to lie in front of the camera, with a margin of $1$m from the camera. For 2D reprojections of generated skeletons, we restrict random camera rotation by uniformly sampling an azimuth angle between $[-\pi, \pi]$ and an elevation angle between $[-\pi/9, \pi/9]$.  \\

We provide supplementary qualitative and quantitative results for our PoseNet3D approach.

\section{Per-activity Evaluation}
In Table~\ref{tbl:pose_results_best_p2}, we present P-MPJPE for each class in Human3.6M dataset. The results shown use the 2D pose detections extracted by SH~\cite{stacked-hourglass} as input. We compare our results with the previous unsupervised approach of Chen~\etal~\cite{chingCVPR2019} and other weakly-supervised methods~\cite{InterpreterNetwork2016, tung_2017_ICCV, ZedNet_2018_ECCVW, wang2019distill}. We outperform previous unsupervised approach of~\cite{chingCVPR2019} on all activity classes. 
\begin{table*}[h]
	    \centering
    	\small
    	\tabcolsep=1mm
    	\resizebox{\textwidth}{!}{%
    		\begin{tabular}{@{}l|rrrrrrrrrrrrrrr|r@{}}
    		& Dir. & Disc. & Eat & Greet & Phone & Photo & Pose & Purch. & Sit & SitD. & Smoke & Wait & Walk & WalkD. & WalkT. & Avg\\
    		\midrule
    		3Dinterp.~\cite{InterpreterNetwork2016}  & 78.6 & 90.8 & 92.5 & 89.4 & 108.9 & 112.4 & 77.1 & 106.7 & 127.4 & 139.0 & 103.4 & 91.4 & 79.1 & - & - & 98.4\\
    		AIGN~\cite{tung_2017_ICCV}  & 77.6 & 91.4 & 89.9 & 88.0 & 107.3 & 110.1 & 75.9 & 107.5 & 124.2 & 137.8 & 102.2 & 90.3 & 78.6 & - & - & 97.2 \\
    		Drover~\etal~\cite{ZedNet_2018_ECCVW} &60.2 & 60.7 & 59.2 & 65.1 & 65.5 & 63.8 & 59.4 & 59.4 & 69.1 & 88.0 & 64.8 & 60.8 & 64.9 & 63.9 & 65.2 & 64.6\\
    		Wang~\etal~\cite{wang2019distill} & 54.7 & 57.7 & 54.8 & 55.8 & 61.6 & 56.3 & 52.7 & 73.7 & 95.5 & 62.3 & 68.5 & 60.8 & 55.5 & 64.0 & 58.0 & 62.1\\
    		  \midrule
    		Chen~\etal~\cite{chingCVPR2019}~$(\dagger)$~(+) & 55.0 &	58.3 & 67.5 & 61.8 & 76.3 &	64.6	& 54.8 & 58.3 &  89.4 & 90.5 & 71.7 & 63.8 & 65.2 &	63.1	& 65.6 & 68\\
    		PoseNet3D-Teacher~$(\dagger)$   &58.9&60.5&67.1&65.1&71.4&61.8&55.4&52.6&90.3&87.5&67.2&64.1&58.7&63.4&60.2&66.6 \\
    		PoseNet3D-Teacher-FT~$(\dagger)$  &52.2&55.0&58.8&59.9&66.3&60.9&53.1&50.9&80.8&85.9&63.3&61.7&57.3&61.7&54.4&62.1 \\
    		PoseNet3D~$(\dagger)$  &\textbf{49.1}&\textbf{52.4}&\textbf{57.5}&\textbf{56.4}&\textbf{63.5}&\textbf{59.5}&\textbf{51.3}&\textbf{48.4}&\textbf{77.1}&\textbf{81.5}&\textbf{60.4}&\textbf{59.6}&\textbf{53.5}&\textbf{59.1}&\textbf{51.4}&\textbf{59.4} \\
    		\bottomrule
    		\end{tabular}
    	}
    	\caption{\textbf{Human3.6M}. Comparison of P-MPJPE for model-free 3D pose estimation. Results obtained using estimated 2D pose by SH/CPM~\cite{stacked-hourglass,cpm}. Best results amongst the unsupervised approaches are shown in bold. $(\dagger)$ using temporal information, (+) using additional data for training.}
    	\label{tbl:pose_results_best_p2}
\end{table*}

\begin{figure*}[t!]
	\centering
	\includegraphics[width=0.99\textwidth]{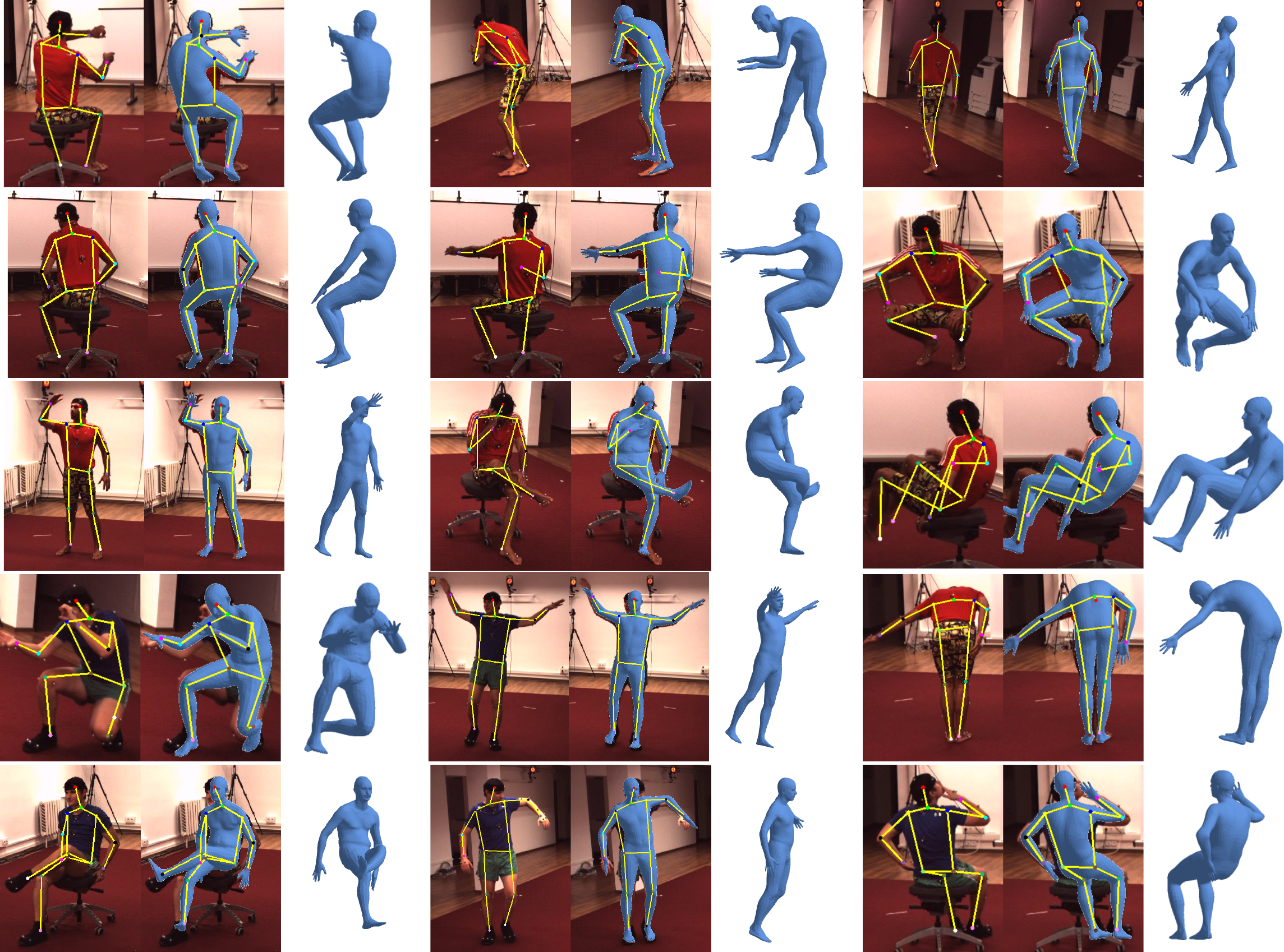} 
	\caption{Visualization of SMPL mesh obtained using predicted parameters on challenging examples from Human3.6M. Each example shows input image with input 2D landmarks, recovered SMPL mesh with reprojected 2d pose predictions, and the same mesh from a different view. }
	\label{fig:h36vis_mesh_more}
\end{figure*}

\begin{figure*}[t!]
	\centering
	\includegraphics[width=0.99\textwidth]{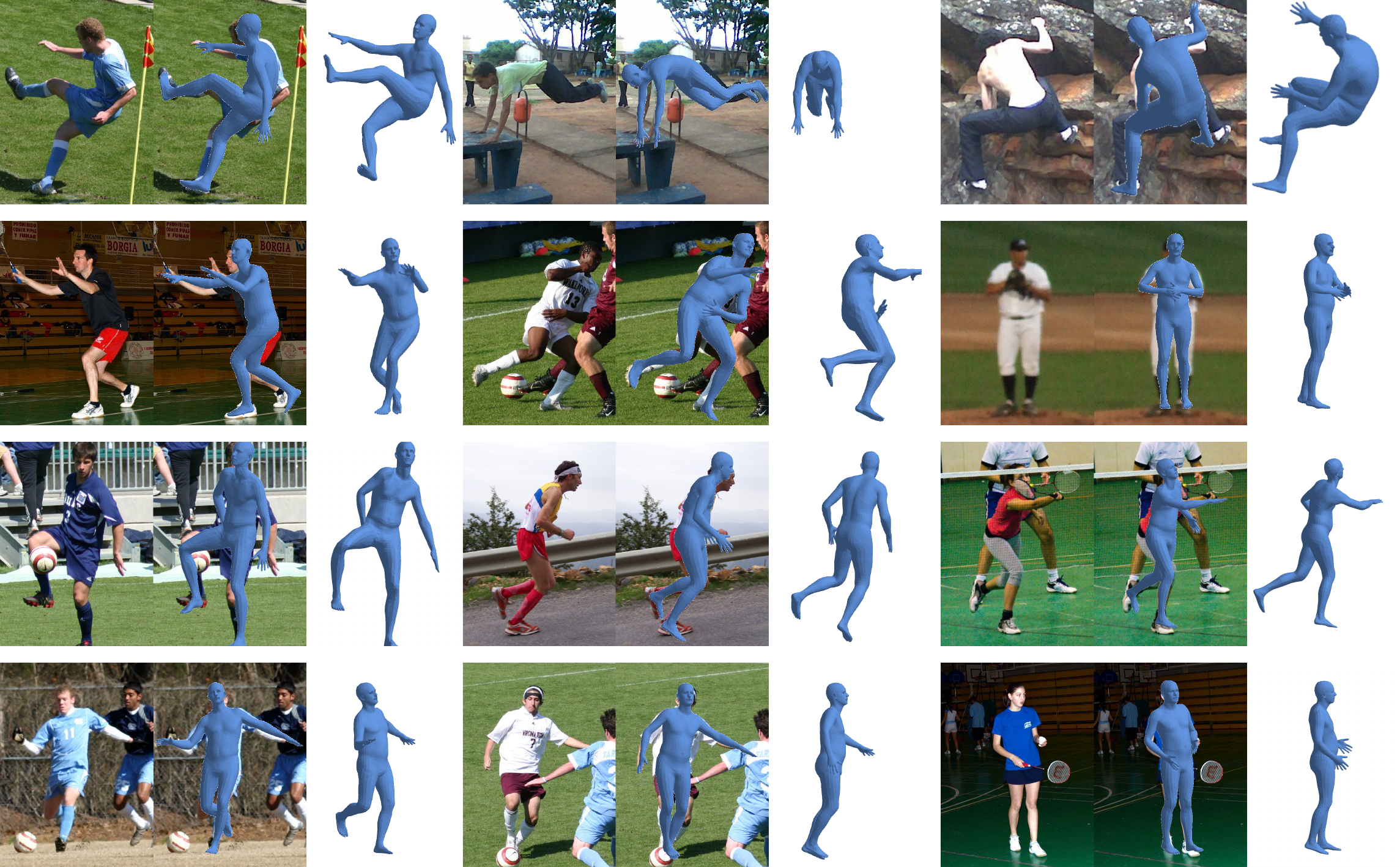} 
	\caption{Visualization of SMPL mesh obtained using predicted parameters on challenging examples from LSP. Each example shows input image, recovered SMPL mesh, and the same mesh from a different view.}
	\label{fig:lsp_vis_mesh_more}
\end{figure*}

\begin{figure*}[t!]
	\centering
	\includegraphics[width=0.99\textwidth]{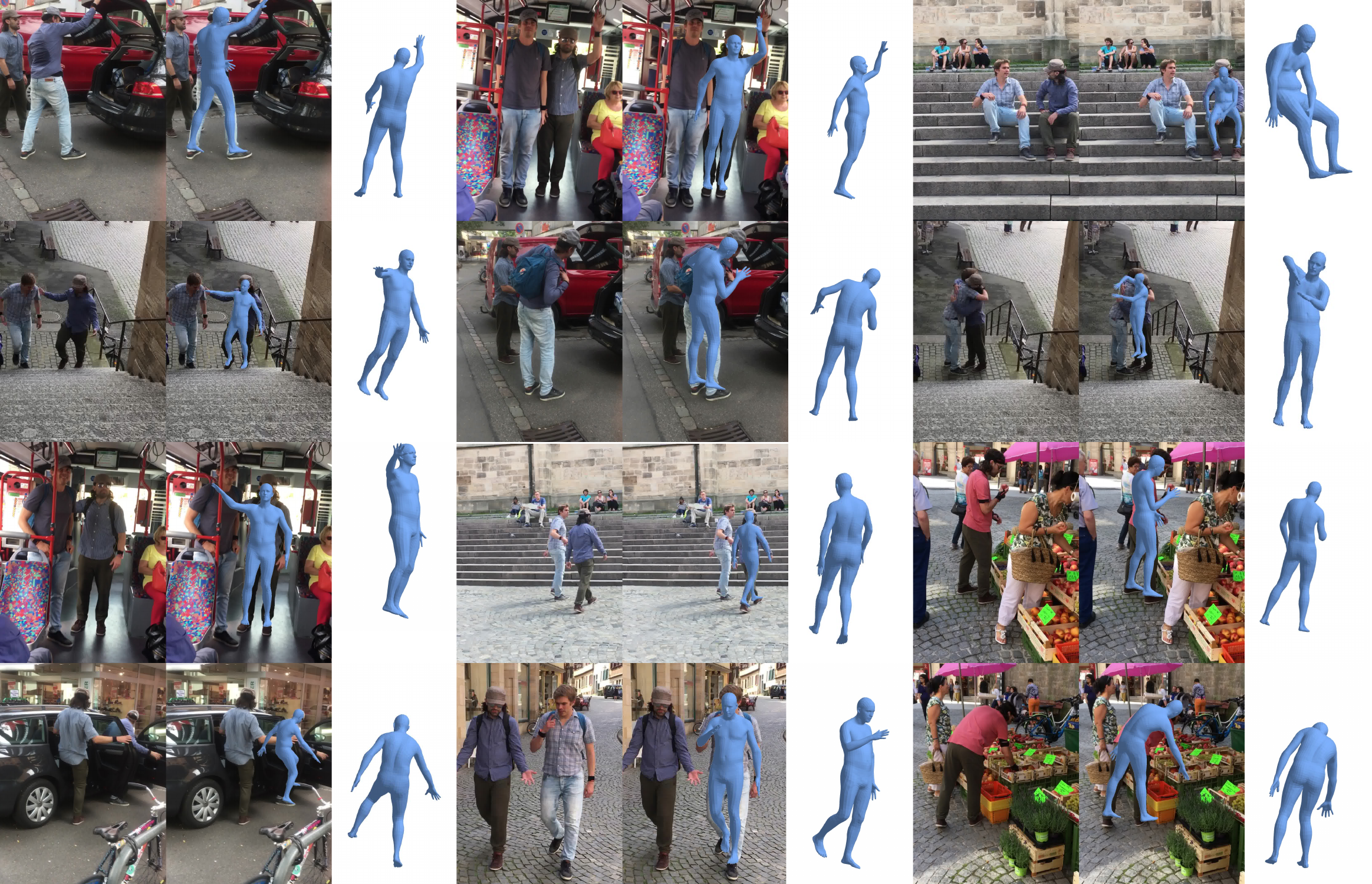} 
	\caption{Visualization of SMPL mesh obtained using predicted parameters on challenging examples from 3DPW. Each example shows input image, recovered SMPL mesh, and the same mesh from a different view.}
	\label{fig:3dpw_vis_supp}
\end{figure*}

\section{Qualitative Results}
Figure~\ref{fig:h36vis_mesh_more}, \ref{fig:lsp_vis_mesh_more} and~\ref{fig:3dpw_vis_supp} show additional visualizations of the predicted SMPL meshes on challenging examples from Human 3.6M, LSP and 3DPW datasets. Note that our approach handles a variety of pose articulations. 

\section{Temporal Evaluation}
\begin{figure*}[t!]
	\centering
	\includegraphics[width=0.99\textwidth]{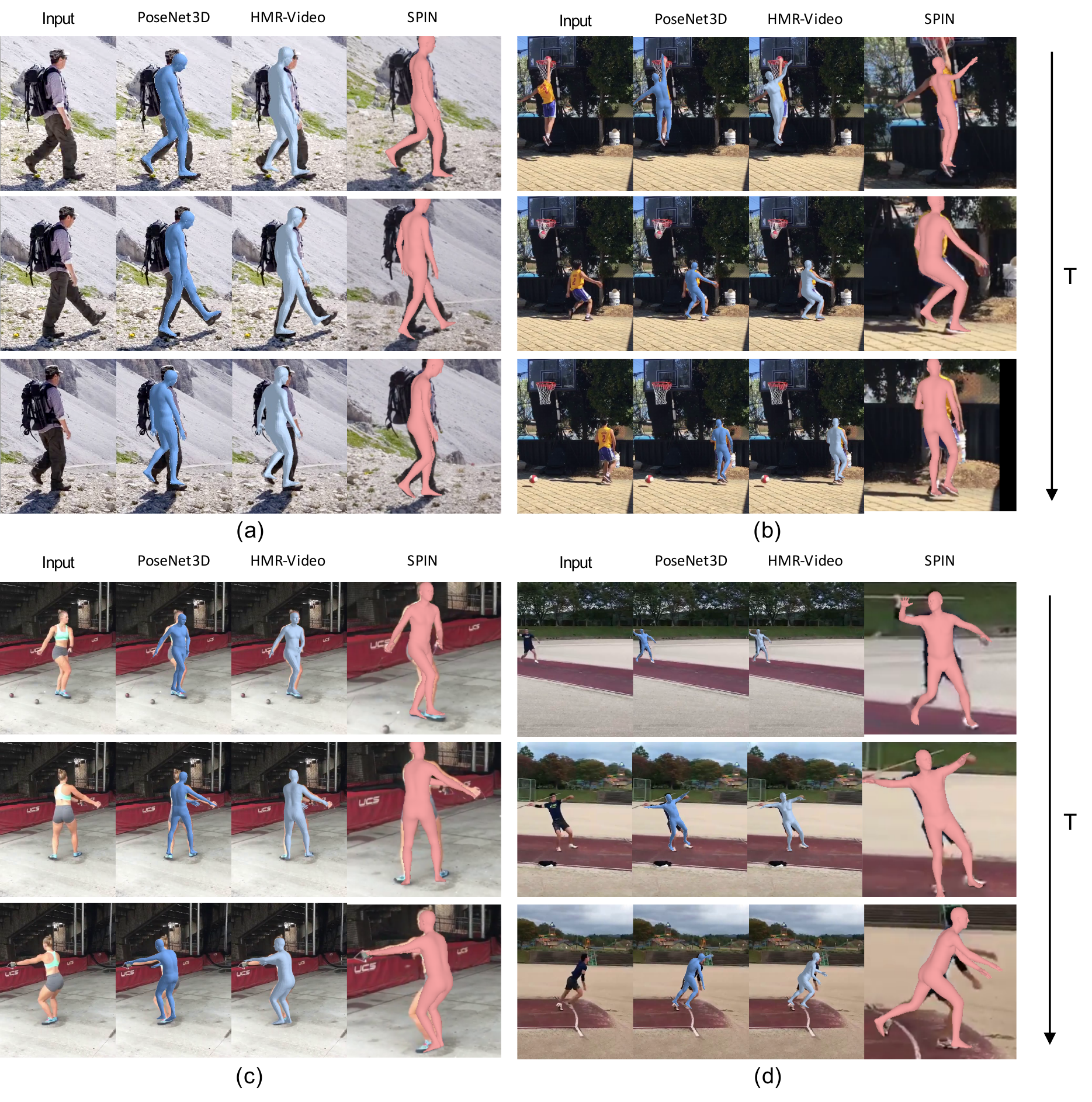} 
	\caption{Qualitative comparison with HMR-Video~\cite{kanazawa2019learning} and SPIN~\cite{SPIN_ICCV2019} on (a) davis-hike (b) insta-variety-dunking (c)  insta-variety-hammerthrow (d) insta-variety-javelinthrow. ($\rightarrow$) indicates the passage of time. }
	\label{fig:temporal_comparison}
\end{figure*}

In Fig.~\ref{fig:temporal_comparison}, we present a side-by-side comparison of the qualitative performance of our approach with two recently published approaches: HMR-Video~\cite{kanazawa2019learning} and SPIN~\cite{SPIN_ICCV2019}. For obtaining the corresponding meshes from~\cite{kanazawa2019learning} and~\cite{SPIN_ICCV2019}, we use their publicly available code and do not post-process their results in any form. 

Please also find video results of our method on temporal sequences \href{https://www.youtube.com/playlist?list=PL46Tof4i1hsHK6D-y8oBSRK-DpmvfyVlf}{here}. We use publicly available in-the-wild videos provided by~\cite{kanazawa2019learning}, PennAction and 3DPW datasets to generate video sequences of our predicted SMPL meshes. It is worth noting that our method is \textit{only trained} on 2D poses from \textit{Human3.6M} dataset, while the results are shown on in-the-wild video sequences, which have no overlap in terms of the action categories between the training and evaluation datasets.

\end{document}